\documentclass{llncs}

\usepackage[belowskip=-17pt,aboveskip=5pt]{caption}

\usepackage{amsmath}
\usepackage{amssymb}

\usepackage[english]{babel}

\allowdisplaybreaks[1]

\usepackage{latexsym}
\usepackage[emu]{cmll}
\usepackage{prooftree}

\def\Section#1{\section{#1}}

\def\Hide#1{\relax}

\DeclareSymbolFont{AMSb}{U}{msb}{m}{n}
\DeclareSymbolFontAlphabet{\mathbb}{AMSb}
\DeclareSymbolFont{symbolsC}{U}{txsyc}{m}{n}
\DeclareMathSymbol{\rJoin}{\mathrel}{symbolsC}{89}

\def\VL{\mathbf{L}}
\def\VM{\mathbf{M}}

\def\Func#1{{\mathsf{#1}}}

\def\iAnd{\otimes} 

\def\Inf{\Func{inf}}

\def\Not{\lnot}

\def\ST{\mathrel{|}}

\def\To{\rightarrow}

\newtheorem{Definition}{Definition}
\newtheorem{Theorem}{Theorem}
\newtheorem{Lemma}[Theorem]{Lemma}

\def\Proof{\par \noindent{\bf Proof: }}
\def\Done{\hfill\rule{0.5em}{0.5em}}
\def\Var{\Func{Var}}

\def\dotplus{\mathop{\dot{+}}}

\def\MONE{[{\sf m}_1]}
\def\MTWO{[{\sf m}_2]}
\def\MTHREE{[{\sf m}_3]}
\def\Mi{[{\sf m}_i]}
\def\OONE{[{\sf o}_1]}
\def\OTWO{[{\sf o}_2]}
\def\OTHREE{[{\sf o}_3]}
\def\OFOUR{[{\sf o}_4]}
\def\Oj{[{\sf o}_j]}
\def\BOUNDED{[{\sf b}]}
\def\RESIDUAL{[{\sf r}]}
\def\ONEAX{[{\sf ann}]}

\def\ASM{[\Func{ASM}]}
\def\IL{{\sf IL}}
\def\CE{[{\iAnd}\Func{E}]}
\def\CI{[{\iAnd}\Func{I}]}

\def\Cwc{[{\sf cwc}]}
\def\Csd{[{\sf csd}]}

\def\CON{[\Func{CON}]}

\def\CSD{[\Func{CSD}]}
\def\CWC{[\Func{CWC}]}
\def\DN{[\Func{DNE}]}

\def\EFQ{[\Func{EFQ}]}

\def\HL{[\Func{HLB}]}
\def\HU{[\Func{HUB}]}

\def\LE{[{\Lolly}\Func{E}]}
\def\LI{[{\Lolly}\Func{I}]}

\def\Lnot{{{}^{\perp}}}
\def\Lolly{\multimap}

\def\rImp{\mathop{\rightarrow}}

\def\EFQ{[\Func{EFQ}]}

\def\Lang#1{{\cal L}_{\mathbf{#1}}}

\def\Li{\Lang{1}}
\def\Lh{\Lang{\frac{1}{2}}}

\def\Logic#1#2{\mbox{{\bf #1}}_{\mbox{\bf #2}}}
\def\ALc{\Logic{AL}{c}}
\def\ALi{\Logic{AL}{i}}
\def\BL{\Logic{BL}{\relax}}
\def\CLc{\Logic{CL}{c}}
\def\CLi{\Logic{CL}{i}}
\def\IL{\Logic{IL}{\relax}}
\def\LLc{\Logic{{\L}L}{c}}
\def\LLi{\Logic{{\L}L}{i}}

\title{(Dual) Hoops Have Unique Halving}

\author{Rob Arthan \& Paulo Oliva}

\institute{Queen Mary University of London \\[1mm]
School of Electronic Engineering and Computer Science \\[1mm]
Mile End Road, London E1 4NS}

\begin{document}
\maketitle

\begin{abstract}

Continuous logic extends the multi-valued {\L}ukasiewicz logic   
by adding a halving operator on propositions.
This extension is designed
to give a more satisfactory model theory for continuous structures.
The semantics of these logics can be given using specialisations
of algebraic structures known as hoops and coops.
As part of an investigation into the metatheory of  propositional
continuous logic, we were indebted to Prover9 for finding proofs
of important algebraic laws.
\end{abstract}

\Section{Introduction}\label{sec:introduction}

(Like its title, this chapter begins with a parenthesis concerning notation.
It is common practice to order truth-values by decreasing logical strength, but
the opposite, or {\em dual}, convention is used in the literature that
motivates the present work. So in this chapter $A \ge B$ means
that $A$ is logically stronger than $B$. Accordingly, in the algebraic
structures we will study, 0 models truth rather than falsehood and
conjunction corresponds to an operation written as addition rather than
multiplication. The halves alluded to in the title would otherwise be square
roots.)

Around 1930, {\L}ukasiewicz and Tarski \cite{lukasiewicz-tarski30}
instigated  the study of logics admitting models in which the
truth values are real numbers drawn from some subset $T$
of the interval $[0, 1]$. In these models, with the notational
conventions discussed above, conjunction is capped addition:
$x \dotplus y = \Inf\{x + y, 1\}$.
Boolean logic is the special case when $T = \{0, 1\}$.
These {\L}ukasiewicz logics have been widely studied, e.g., as instances of fuzzy logics \cite{Hajek98}.

In recent years, Ben Yaacov has used a Lukasiewicz logic with an infinite number
of truth values as a building block in what is called continuous logic
\cite{ben-yaacov-pedersen09}. Continuous logic unifies work of Henson and
others \cite{henson-iovino02} that aims to overcome shortfalls of classical
first-order model theory when applied to continuous structures such as metric
spaces and Banach spaces.
A detailed discussion of these shortfalls would be out of place here, but
a few remarks are in order.
In functional analysis there is a well-accepted
notion of ultraproduct that takes into account metric structure and
is an important tool for constructing Banach spaces. By contrast,
the class of Banach spaces is not closed under
the standard model-theoretic notion of ultraproduct.
Continuous logic aims to capture properties that are preserved under the good
notion of ultraproduct for continuous structures \cite{henson-iovino02}.
From another point of view, continuous logic mitigates
the fact that ordinary first-order
logic for continuous structures tends to be unexpectedly strong,
the first-order theory of Banach spaces being strictly
stronger than second-order arithmetic \cite{Solovay-et-al12}.

The motivation for ordering truth values by increasing logical
strength in continuous logic stems from the fact that
in a metric space with metric $d$, $x = y$ iff $d(x, y) = 0$.  In first-order
continuous logic, one wishes to treat $d$ as a two-place
predicate symbol analagous to equality in classical first-order logic.
Representing truth by 0 is then the natural choice.

A difficulty with both the {\L}ukasiewicz logics and continuous logic is that
it requires considerable ingenuity to work with
the known axiomatisations of their propositional fragments.
Work on algebraic semantics for {\L}ukasiewicz logic begun by Chang
\cite{chang58b,chang59} has helped greatly with this, but basic algebraic
laws in the algebras involved are often quite difficult to prove.
This chapter reports on ongoing work to gain a better understanding of both the
proof theory and the semantics of continuous
logic that is benefitting from the use of automated theorem proving
to find counterexamples and to derive algebraic properties.

\setlength{\unitlength}{3mm}
\begin{figure}[h]
\begin{center}
\begin{picture}(23,12)(4,10)
\put(4,7){\makebox(3,2){$\ALi$}}
\put(4,17){\makebox(3,2){$\ALc$}}
\put(14,7){\makebox(3,2){$\LLi$}}
\put(14,17){\makebox(3,2){$\LLc$}}
\put(20,12){\makebox(3,2){$\IL$}}
\put(20,22){\makebox(3,2){$\BL$}}
\put(24,7){\makebox(3,2){$\CLi$}}
\put(24,17){\makebox(3,2){$\CLc$}}
\put(5.5,9.5){\vector(0,1){7.5}}
\put(15.5,9.5){\vector(0,1){7.5}}
\put(25.5,9.5){\vector(0,1){7.5}}
\put(21.5,14.5){\line(0,1){3.25}}
\put(21.5,18.75){\vector(0,1){3.25}}
\put(7.5,8.25){\vector(1,0){6}}
\put(7.5,18.25){\vector(1,0){6}}
\put(17.5,8.25){\vector(1,0){6}}
\put(17.5,18.25){\vector(1,0){6}}
\put(17,9){\vector(1,1){3.25}}
\put(17,19){\vector(1,1){3.25}}
\end{picture}
\vspace{8mm}
\caption{Eight Logics and the Relationships between Them}
\label{fig:logics}
\end{center}
\end{figure}
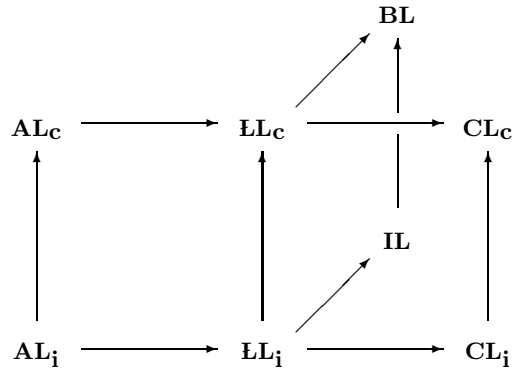

Our work began with the observation that both {\L}ukasiewicz logic, $\LLc$,
and Ben Yaacov's continuous logic, $\CLc$, are extensions of a very simple intuitionistic
substructural logic $\ALi$.  In Section~\ref{sec:the-logics} of this chapter we
show how $\CLc$ may be built up via a system of extensions of $\ALi$. We also
show how the Brouwer-Heyting intuitionistic propositional logic, $\IL$, and Boolean
logic, $\BL$, fit into this picture. The relationships between the eight logics
in this system of extensions are depicted
in Figure~\ref{fig:logics}.
In Section~\ref{sec:algebraic-semantics}, we describe a class of monoids
called pocrims
that have been quite widely studied in connection with $\ALi$ and sketch a
proof of a theorem asserting that each of the eight logics is sound and
complete with respect to an appropriate class of pocrims.  The sketch is easy to
complete apart from one tricky lemma concernng the continuous logics.

In Section~\ref{sec:proofs-and-counterexamples}, we discuss our use of Bill McCune's
Mace4 and Prover9 to assist in these investigations, in particular to
prove the lemma needed for the theorem of Section~\ref{sec:the-logics}.
Our application seems to be a ``sweet spot'' for this kind of technology:
the automatic theorem prover found a proof of a difficult problem that
can readily be translated into a human readable form.

In Section~\ref{sec:subsequent-work} we discuss some other results that
Prover9 has proved for us.
Section~\ref{sec:final-remarks} gives some concluding remarks.

\Section{The Logics}\label{sec:the-logics}

We work in a language $\Lh$ whose
atomic formulas are the propositional constants
$0$ (truth) and $1$ (falsehood) and
propositional variables drawn from
the set $\Var = \{P, Q, \ldots\}$.
If $A$ and $B$ are formulas of $\Lh$ then so
are $A \Lolly B$ (implication),
$A \iAnd B$ (conjunction) and  $A/2$ (halving).
We adopt the convention that implication associates
to the right and has lower precedence than conjunction,
which in turn has lower precedence than halving. So, for example,
the brackets in $(A \iAnd  (B/2)) \Lolly (C \Lolly (D \iAnd F))$ are all redundant, while
those in $(((A \rImp B) \rImp C) + D)/2$ are all required.
We denote by $\Li$ the language without halving. 
We write $A\Lnot$ as an abbreviation for $A \Lolly 1$, a form of negation.

The judgements of the eight logics that we will consider are sequents
$\Gamma \vdash A$, where $A$ is an  $\Lh$-formula
and $\Gamma$ is a multiset of $\Lh$-formulas.
The inference rules are the introduction and elimination rules for the two
binary connectives\footnote{Omitting disjunction from the logic
greatly simplifies the algebraic semantics. While it may be
unsatisfactory from the point of view of intuitionistic philosophy,
disjunction defined using de Morgan's law is adequate for our purposes.}
shown in Figure~\ref{fig:rules}.

\begin{figure}[!t]
\centering
\begin{tabular}{|ccc|}
\hline
\ & \quad \quad & \\
%
%
\begin{prooftree}
\Gamma, A \vdash B
\justifies
\Gamma \vdash A \Lolly B
\using{\LI}
\end{prooftree} & &
%
%
\begin{prooftree}
\Gamma \vdash A \quad \Delta \vdash A \Lolly B
\justifies
\Gamma, \Delta \vdash B
\using{\LE}
\end{prooftree} \\
\ & & \\
%
%
\; \begin{prooftree}
\Gamma \vdash A \quad \Delta \vdash B
\justifies
\Gamma, \Delta \vdash A \iAnd B
\using{\CI}
\end{prooftree} & &
%
%
\begin{prooftree}
\Gamma \vdash A \iAnd B \quad \Delta, A, B \vdash C
\justifies
\Gamma, \Delta \vdash C
\using{\CE}
\end{prooftree} \; \\
\ & & \\
\hline
\end{tabular}
\caption{The Inference Rules}
\label{fig:rules}
\end{figure}

The axiom schemata for the logics are selected from those shown in Figure~\ref{fig:axioms}.
These are the axiom of assumption $\ASM$, ex-falso-quodlibet $\EFQ$, double negation elimination $\DN$, commutative weak conjunction $\CWC$, commutative strong disjunction $\CSD$, the axiom of contraction $\CON$,
and two axioms giving lower and upper bounds for the halving operator:
$\HL$ and $\HU$.

$\ASM, \EFQ$, $\DN$ and $\CON$ are standard axioms of classical logic.  $\CON$
asserts that $A$ is a strong as $A \iAnd A$ and is equivalent to the rule of
{\em contraction} allowing us to infer $\Gamma, A \vdash B$, from $\Gamma, A, A
\vdash B$. $\CON$ allows one to think of the contexts $\Gamma$ as sets rather
than multisets.  The significance of $\CWC$, $\CSD$, $\HL$ and $\HU$ will be
explained below as we introduce the logics that include them and as we give the
semantics for those logics.

The definitions of the eight logics are discussed in the next few paragraphs
and are summarised in Table~\ref{tab:models}.  In all but $\CLi$ and $\CLc$,
halving plays no r\^{o}le and the logical language may be taken to be the
sublanguage $\Li$ in which halving does not feature.

\begin{figure}[!t]
\centering
\begin{tabular}{|cc|}\hline
\ & \\
%
%
\begin{prooftree}
\justifies
\Gamma, A \vdash A
\using{\ASM}
\end{prooftree} &
%
%
\begin{prooftree}
\justifies
\mathstrut \Gamma, 1 \vdash A
\using{\EFQ}
\end{prooftree}
\\\ &  \\
%
%
\begin{prooftree}
\justifies
\Gamma, A\Lnot\Lnot \vdash A
\using{\DN}
\end{prooftree} &
%
%
\begin{prooftree}
\justifies
\Gamma, A \iAnd (A \Lolly B) \vdash B \iAnd (B \Lolly A)
\using{\CWC}
\end{prooftree} \;
\\\ &  \\
%
%
\; \begin{prooftree}
\justifies
\Gamma, (A \Lolly B) \Lolly B \vdash (B \Lolly A) \Lolly A
\using{\CSD}
\end{prooftree} &
%
%
\begin{prooftree}
\justifies
\Gamma, A \vdash A \iAnd A
\using{\CON}
\end{prooftree}
\\\ & \\
%
%
\begin{prooftree}
\justifies
\Gamma, A/2, A/2 \vdash A
\using{\HL}
\end{prooftree} &
%
%
\begin{prooftree}
\justifies
\Gamma, A/2 \Lolly A \vdash A/2
\using{\HU}
\end{prooftree}
\\\ & \\
\hline
\end{tabular}
\caption{The Axiom Schemata}
\label{fig:axioms}
\end{figure}

{\em Intuitionistic affine logic} \cite{Bierman(1993)}, $\ALi$, has for its axiom schemata $\ASM$ and $\EFQ$. All our other logics include $\ALi$.
The contexts $\Gamma$, $\Delta$ are multisets because we wish to keep track of how many times each of the
assumptions in $\Gamma$ is used in order to derive the conclusion $A$ in $\Gamma \vdash A$. This is not relevant if formulas can be duplicated or contracted (i.e. if $A$ is equivalent to $A \iAnd A$).
We will, however, mainly work with so-called {\em substructural logics} where such equivalences are not valid in general. $\ALi$ serves as a prototype for such substructural logics. 

Under the Curry-Howard correspondence between proofs and $\lambda$-terms, the proof system $\ALi$ corresponds to a $\lambda$-calculus with
pairing and paired abstractions, so in this calculus, if $t$, $u$ and $v$ are terms,
then so are $(t, u)$, $(t, (u, v))$,$\lambda (x, y)\bullet t$,
$\lambda((x, y), z)\bullet u$, $\lambda(x, (y, z))\bullet v$ etc. where
$x$, $y$ and $z$ are variables. Proofs in $\ALi$ then correspond to
{\em affine} $\lambda$-terms: terms
in which each variable is used at most once.
So for example $\lambda f \bullet \lambda x \bullet \lambda y \bullet f(x, y)$
is an affine $\lambda$-term corresponding to a proof of
the sequent $\vdash (A \iAnd B \Lolly C) \Lolly A \Lolly B \Lolly C$.

\begin{table}
\centering
\begin{tabular}{|c|l|l|} \hline
{\bf Logic} & {\bf Axioms} & {\bf Models} \\
\hline \hline
$\ALi$ & $\ASM + \EFQ$ & bounded pocrims \\[1mm]
\hline
$\ALc$ & $\ALi + \DN$& bounded involutive pocrims \\[1mm]
\hline
$\LLi$ & $\ALi + \CWC$ & bounded hoops \\[1mm]
\hline
$\LLc$ & $\ALi + \CSD$ & bounded Wajsberg hoops \\[1mm]
\hline
$\IL$  & $\ALi + \CON$ & bounded idempotent pocrims \\[1mm]
\hline
$\BL$  & $\IL + \DN$ & bounded involutive idempotent pocrims \\[1mm]
\hline
$\CLi$ & $\LLi + \HL + \HU$ & bounded coops \\[1mm]
\hline
$\CLc$ & $\LLc + \HL + \HU$ & bounded involutive coops \\[1mm]
\hline
\end{tabular}
\vspace{4mm}
\caption{The Logics and their Models}
\vspace{-8mm}
\label{tab:models}
\end{table}

{\em Classical affine logic} \cite{Girard(87B)}, $\ALc$, extends $\ALi$ with the
axiom schema $\DN$. It can also be viewed as the extension of the so-called
multiplicative fragment of Girard's linear logic by allowing weakening and the
axiom schema $\EFQ$.

What we will call {\em intuitonistic {\L}ukasiewicz logic},
$\LLi$, extends $\ALi$ with the axiom schema $\CWC$.
$\LLi$ is known by a variety of names in the literature.
The name we use reflects its position in Figure~\ref{fig:logics}.
For any formulas $A$ and $B$,
$A \iAnd (A \Lolly B)$ implies both $A$ and $B$ and
so can be thought of as a weak form of conjunction.
In $\LLi$ we have commutativity of this weak conjunction.
$\CWC$ turns out to be a surprisingly powerful axiom.
However, it often requires considerable ingenuity to use it.

{\em Classical {\L}ukasiewicz logic} \cite{Hay(1963)}, $\LLc$, extends $\ALi$ with the axiom
schema $\CSD$. Just as $A \iAnd (A \Lolly B)$ can be viewed
as a form of conjunction, $(A \Lolly B) \Lolly B$ can be viewed as
a form of disjunction that may be stronger than the one defined by
the usual intuitionistic rules for disjunction.
In $\LLc$ we have commutativity of this strong disjunction.
This gives the widely-studied multi-valued logic of {\L}ukasiewicz.
Like $\CWC$, $\CSD$ is powerful but not always easy to use.

{\em Intuitionistic propositional logic}, $\IL$, extends $\ALi$ with the axiom
schema of contraction $\CON$. This gives us the
conjunction-implication fragment of the well-known Brouwer-Heyting
intuitionistic propositional logic.

{\em Classical propositional logic} (or boolean logic), $\BL$, extends $\IL$ with the axiom schema $\DN$.
This is the familiar two-valued logic of truth tables.

What we have termed {\em intuitionistic continuous logic}, $\CLi$, allows the
halving operator and extends $\LLi$ with the axiom schemas $\HL$ and $\HU$,
which effectively give lower and upper bounds on the logical strength of $A/2$.
They imply the surprisingly strong condition that $A/2$ is equivalent to $A/2
\Lolly A$. This is an intuitionistic version of the continuous logic of Ben
Yaacov \cite{ben-yaacov-pedersen09}.

{\em Classical continuous logic}, $\CLc$ extends $\CLi$ with the axiom
schema $\DN$. This gives Ben Yaacov's continuous logic. The motivating
model takes truth values to be real numbers between $0$ and $1$ with
conjunction defined as capped addition.

Our initial goal was to gain insight into $\CLc$ by investigating the relations
amongst $\ALi$, $\LLc$ and $\CLc$. The other logics came into focus when we
tried to decompose the somewhat intractable axiom $\CSD$ into a combination of
$\DN$ and an intuitionistic component.  It can be shown that the eight logics
are related as shown in Figure~\ref{fig:logics}. In the figure, an arrow from
$T_1$ to $T_2$ means that $T_2$ extends $T_1$, i.e., the set of provable
sequents of $T_2$ contains that of $T_1$. In each square, the north-east logic
is the least extension of the south-west logic that contains the other two. For
human beings, at least, the proof of this fact is quite tricky for the
$\ALi$-$\LLc$ square, see~\cite[chapters 2 and 3]{Hajek98}.

The routes in Figure~\ref{fig:logics} from $\ALi$ to $\IL$ and $\BL$
have been quite extensively studied, as may be seen from
\cite{blok-ferreirim00,raftery07,koehler81} and the works cited therein.
We are not aware of any work on $\CLi$, but it is clearly a natural
object of study in connection with Ben Yaacov's continuous logic.
It should be noted that $\IL$ and $\CLi$ are incompatible: as we
will see at the end of this section, any formula is provable given the
axioms $\CON$, $\HL$ and $\HU$.


\Section{Algebraic Semantics}\label{sec:algebraic-semantics}

We give algebraic semantics for the logics of Section~\ref{sec:the-logics}
using {\em pocrims}:
partially ordered, commutative, residuated, integral monoids.

\begin{Definition} A {\em pocrim}\footnote{Strictly speaking, this is a {\em dual} pocrim,
since we order it by increasing logical strength and write it additively.}
is a structure for the signature $(0, +, \rImp; \ge)$ of type
$(0, 2, 2; 2)$ satisfying the following laws:

\[
\begin{array}{l@{\quad\quad}r}
(x + y) + z = x + (y + z) & \MONE \\
x + y = y + x & \MTWO \\
x + 0 = x & \MTHREE \\
x \ge x & \OONE \\
\mbox{if $x \ge y$ and $y \ge z$, then $x \ge z$} & \OTWO \\
\mbox{if $x \ge y$ and $y \ge x$, then $x = y$} & \OTHREE \\
\mbox{if $x \ge y$, then $x + z \ge y + z$} & \OFOUR \\
x \ge 0 & \BOUNDED \\
\mbox{$x + y \ge z$ iff $x \ge y \rImp z$} & \RESIDUAL
\end{array}
\]
\end{Definition}

Intuitively, $\rImp$ is the semantic counterpart of the syntactic implication $\Lolly$, whereas $+$ corresponds to the syntactic conjunction $\iAnd$. 
As with the syntactic connectives,,
we adopt the convention that $\rImp$ associates
to the right and has lower precedence than $+$.
Note that $=$ and $\ge$ are predicate symbols and so necessarily have
lower precedence than the function symbols $\rImp$ and $+$: the
only valid reading of $a \rImp b \ge c + d$ is as $(a \rImp b) \ge (c + d)$.

Let $\VM = (M, 0, +, \rImp; \ge)$ be a pocrim.
The laws~$\Mi$, $\Oj$ and $\BOUNDED$ say that
$(M, 0, +; {\ge})$ is a partially ordered commutative monoid
with the identity $0$ as least element.
Law~$\RESIDUAL$, the {\em residuation property},
says that for any $x$ and $z$ the
set $\{y \ST x + y \ge z\}$ is non-empty and has $x \rImp z$
as least element.
$\VM$ is said to be {\em bounded} if it has a (necessarily unique)
{\em annihilator}, i.e., an element $1$ such that for every $x$ we have:
\[
\begin{array}{l@{\quad\quad}r}
1 = x + 1 & \ONEAX
\end{array}
\]

Let us assume $\VM$ is bounded. Then $1 = x + 1 \geq x \geq 0$ for any $x$ and
$(M; \ge)$ is indeed a bounded ordered set.
Let $\alpha : \Var \To M$ be an interpretation of logical variables
as elements of $M$
and extend $\alpha$ to a function $v_{\alpha} : \Li \To M$
by interpreting $0$, $1$, $\iAnd$ and $\Lolly$ as $0$, $1$, $+$ and
$\rImp$ respectively. If $\Gamma = C_1, \ldots, C_n$,
we say that $\alpha$ {\em satisfies} the sequent
$\Gamma \vdash A$, iff $v_{\alpha}(C_1) + \ldots + v_{\alpha}(C_n) \ge v_{\alpha}(A)$.
We say that $\Gamma \vdash A$ is {\em valid in} $\VM$
if it is satisfied by every assignment $\alpha : \Var \To M$.
We say $\VM$ is a {\em model} for a logic $\VL$ if every sequent provable in
$\VL$ is valid in $\VM$.
If $\cal C$ is a class of pocrims, we say $\Gamma \vdash A$ is {\em valid}
if it is valid in every $\VM \in {\cal C}$.

We will need some special classes of pocrim.
We write $\Not x$ as an abbreviation for $x \rImp 1$, a semantic
analogue of the derived syntactic operator $\Lnot$.
We say a bounded pocrim is {\em involutive} if it 
satisfies $\Not\Not x = x$.
We say a pocrim is {\em idempotent}
if it is idempotent as a monoid, i.e., it satisfies $x + x = x$.

\begin{Definition}[B\"{u}chi \& Owens\cite{BO}] A {\em hoop}\footnote%
{
B\"{u}chi and Owens \cite{BO} write of hoops that ``their importance \ldots
merits recognition with a more euphonious name than the merely descriptive
``commutative complemented monoid''''. Presumably they chose ``hoop'' as a
euphonious companion to ``group'' and ``loop''.
}
is a pocrim that is {\em naturally ordered}, i.e., whenever $x \ge y$, there is
$z$ such that $x = y + z$. 
\end{Definition}

It is a nice exercise in  the use of the residuation
property to show that a pocrim is a hoop iff it satisfies the identity
\[
\begin{array}{l@{\quad\quad}r}
x + (x \rImp y) = y + (y \rImp x) & \Cwc
\end{array}
\]
In any pocrim, $x \le x + (x \rImp y) \ge y$, so we can view $x + (x \rImp y)$
as a weak form of conjunction, but in general this conjunction is not
commutative and there need be no least $z$ such that $x \le z \ge y$.  In a
hoop, the weak conjunction is commutative and $x + (x \rImp y)$ can be shown to
be the least upper bound of $x$ and $y$.

\begin{Definition}[Blok \& Ferreirim\cite{blok-ferreirim00}]
A {\em Wajsberg hoop} is a hoop satisfying the identity
\[
\begin{array}{l@{\quad\quad}r}
(x \rImp y) \rImp y = (y \rImp x) \rImp x & \Csd
\end{array}
\]
\end{Definition}
We may view $(x \rImp y) \rImp y$ as a form of disjunction. In a
Wajsberg hoop this disjunction is commutative and can be shown
to give a greatest lower bound of $x$ and $y$.
See~\cite{blok-ferreirim00} for more information on hoops and Wajsberg hoops.

\begin{Definition}
A {\em continuous hoop}, or {\em coop}, is a hoop where for every $x$ there is a unique $y$
such that $y = y \rImp x$. In this case we write $y = x/2$.
\end{Definition}
In a coop, for any $x$, we have $x \ge x/2 \rImp x = x/2$, whence, by $\Cwc$,
$x = x + 0 = x + (x \rImp x/2) = x/2 + (x/2 \rImp x) = x/2 + x/2$, justifying
our choice of notation. Here, as with the syntactic connectives, we take halving
to have higher precedence than conjunction.

If $\VM$ is a coop, we extend the function $v_{\alpha} : \Li \To M$
induced by an interpration $\alpha :\Var \To M$ to a function
$v_{\alpha} : \Lh \To M$ by interpreting $A/2$ as $v_{\alpha}(A)/2$.
The notions of validity and satisfaction extend to interpretations of $\Lh$
in a coop in the evident way.

We say that a logic $L$ is sound for a class of pocrims $\cal C$
if every sequent that is provable in $L$ is valid in $\cal C$.
We say that $L$ is complete for $\cal C$ if the converse holds.
We then have:

\begin{Theorem}\label{thm:ali-sound-complete}
Each of the logics
$\ALi$, $\ALc$, $\LLi$, $\LLc$, $\IL$, $\BL$, $\CLi$ and $\CLc$
is sound and complete for the class of pocrims listed for it
in the column headed ``Models'' in Table~\ref{tab:models}.
\end{Theorem}
\Proof
The proof follows a standard pattern and, with one exception,
filling in the details is straightforward.
Soundness is a routine exercise. For the completeness, one defines
an equivalence relation $\simeq$ on formulas such that $A \simeq B$ holds
iff both $A \vdash B$ and $B \vdash A$ are provable in the logic.
One then shows that the set of equivalence classes becomes a pocrim
in the indicated class, the {\em term model}, under operators $+$ and $\rImp$
induced on the equivalence classes by $\iAnd$ and $\Lolly$.  As the only
sentences valid in the term model are those provable in the logic, completeness
follows. The difficult detail is showing that the
term models for the continuous logics satisfy our definition of a coop:
it is easy to see that for any $x = [A]$, one has that $y = [A/2]$ satisfies
$y = y \rImp x$, but is this $y$ unique? We shall answer this question
in the affirmative in the next section. If the equation $y = y \rImp x$
did not uniquely determine $y$, halving would not be well-defined
on the term model and the completeness proof would fail.
\Done

\vspace{2mm}

Using Theorem~\ref{thm:ali-sound-complete}, we can give an algebraic proof of
the claim made earlier that $\IL$ and $\CLi$ are incompatible. By dint of the
theorem, this is equivalent to the claim that a bounded idempotent coop is the
trivial coop $\{0\}$. We may prove this as follows: if $a$ is an element of a
coop and $a/2$ is idempotent, so that $a/2 = a/2 + a/2$, then $a/2 \ge a/2 +
a/2 = a$, so by the residuation property, $a/2 \rImp a = 0$. Now $a/2 = a/2
\rImp a$ by the definition of a coop, so we have $a = a/2 + a/2 = (a/2 \rImp a)
+ (a/2 \rImp a) = 0 + 0 = 0$.

\Section{Automated Proofs and Counterexamples}\label{sec:proofs-and-counterexamples}

In our early attempts to understand the relationships represented in
Figure~\ref{fig:logics}, we spent some time devising finite pocrims
with interesting properties. This can be a surprisingly difficult and
error-prone task. Verifying associativity, in particular, is irksome. Having
painstakingly accumulated a small stock of examples, a conversation with Alison
Pease reminded us of the existence of Bill McCune's Mace4
tool~\cite{prover9-mace4} that automatically searches for finite
counter-examples to conjectures in a finitely axiomatised first-order theory.

It was fascinating to see Mace4 recreate examples similar to those
we had already constructed. The following input asks Mace4 to produce
a counterexample to the conjecture that all bounded pocrims are hoops:

\begin{center}
\begin{verbatim}
op(500, infix, "==>").
formulas(assumptions).
   (x + y) + z = x + (y + z).       % monoid law 1
   x + y = y + x.                   % monoid law 2
   x + 0 = x.                       % monoid law 3
   x >= x.                          % ordering law 1
   x >= y & y >= z -> x >= z.       % ordering law 2
   x >= y & y >= x -> x = y.        % ordering law 3
   x >= y -> x + z >= y + z.        % ordering law 4
   x >= 0.                          % boundedness law
   x + 1 = 1.                       % annihilator law
   x + y >= z <-> x >= y ==> z.     % residuation law
end_of_list.
formulas(goals).
   x + (x ==> y) = y + (y ==> x).   % can we derive cwc?
end_of_list.
\end{verbatim}
\end{center}

Here we use
`\verb|==>|' and `\verb|>=|' to represent `$\rImp$' and `$\ge$'
in the pocrim and `\verb|&|', `\verb|->|' and `\verb|<->|'
are Mace4 syntax for logical conjunction, implication and bi-implication.
Given the above, Mace4 quickly prints out the diagram
of a pocrim on the ordered set  $0 < p < q < 1$ with
$x + y = 1$ whenever $\{x, y\} \subseteq \{p, q, 1\}$, a counter-example which we had already
come up with over the course of an afternoon. That led us to test Mace4 on
yet other conjectures which we had already refuted with some small counter-examples.
Mace4, again and again, came up with similar counter-models to the ones we had contrived.

Some weeks later we wanted to show that the two axiom schemata $\HL$
and $\HU$ uniquely determine the halving operator over the logic $\LLi$, which would conclude the proof of Theorem \ref{thm:ali-sound-complete}. That would give
us an intuitionistic counterpart ($\CLi$) to continuous logic $\CLc$. In logical terms, we wanted to show that the rule shown in Figure~\ref{fig:half-equals-rule} is derivable in $\LLi$:
\begin{figure}
\centering
\begin{tabular}{|ccc|}\hline
\ &&\\
$\quad$ &
\begin{prooftree}
A \Lolly B \vdash A \quad \quad A \vdash A \Lolly B \quad \quad  C \vdash C \Lolly B \quad \quad C \Lolly B \vdash C
\justifies
A \vdash  C
\end{prooftree} &
$\quad$
\\\ &&\\\hline
\end{tabular}
\caption{A Conjectured Inference Rule}
\label{fig:half-equals-rule}
\end{figure}

After several failed attempts to find a proof, we had
started to wonder whether the rule was not derivable.  That
is when we thought of using Prover9 to look for a proof. We gave Prover9
the input shown below comprising the laws
for a hoop, the assumptions $a
\rImp b = a$ (corresponding to $A \Lolly B \vdash A$ and $A \vdash A \Lolly B$)
and $c \rImp b = c$ (corresponding to $C \vdash C \Lolly B$ and $C \Lolly B
\vdash C$) and the goal $a = c$.
(Because the conjectured inference rule is symmetric in $A$ and $C$,
if the rule is valid, then the antecedents imply that $A$ and $C$ are equivalent).

\begin{center}
\begin{verbatim}
op(500, infix, "==>").
formulas(assumptions).
   (x + y) + z = x + (y + z).       % monoid law 1
   x + y = y + x.                   % monoid law 2
   x + 0 = x.                       % monoid law 3
   x >= x.                          % ordering law 1
   x >= y & y >= z -> x >= z.       % ordering law 2
   x >= y & y >= x -> x = y.        % ordering law 3
   x >= y -> x + z >= y + z.        % ordering law 4
   x >= 0.                          % boundedness law
   x + y >= z <-> x >= y ==> z.     % residuation law
   x + (x ==> y) = y + (y ==> x).   % cwc
   a ==> b = a.                     % assumption 1
   c ==> b = c.                     % assumption 2
end_of_list.
formulas(goals).
   a = c.
end_of_list.
\end{verbatim}
\end{center}

To our surprise Prover9 took just a few seconds to produce the proof shown in the appendix.
The proof that Prover9 found seems perplexingly intricate at first glance, but
after studying it for a little while, we found we could edit it into a form fit
for human consumption.  From a human perspective, the proof involves the 9
intermediate claims given in the following lemma. Once these are proved, we
will see that the desired result is an easy consequence of claim (9),


%

\begin{Lemma} \label{claims}
Let $\VM = (M, 0, +, \rImp; \ge)$ be a hoop and let $a, b, c, x, y \in M$.
Assume that, ($i$), $a \rImp b = a$ and,
($ii$), $c \rImp b = c$. Then the following hold:
\begin{align*}
	\setlength{\itemsep}{4pt}
	(1) \quad & b \geq a \mbox{ and } b \geq c, \\
	(2) \quad & a + a = b, \\
	(3) \quad & a \rImp (a \rImp c) = 0, \\
	(4) \quad & (x \rImp y) + z \geq x \rImp (y + (y \rImp x) + z), \\
	(5) \quad & c \rImp (a + a + x) \geq c, \\
	(6) \quad & c \rImp a \geq a \rImp c, \\
	(7) \quad & c \rImp a = a \rImp c, \\
	(8) \quad & c + (c \rImp a) + ((a \rImp c) \rImp a) = b, \\
	(9) \quad & a + c = b.
\end{align*}
\end{Lemma}
\Proof
In the proof below (in)equalities which are not labelled as following
from one of the assumptions
($i$) and ($ii)$ or an earlier part of the lemma
follow immediately from the axioms of a pocrim. \\[1mm]
\noindent(1) 
We have $b \geq a \rImp b$ and, by ($i$), $a \rImp b = a$).
So $b \geq a$ and similarly $b \geq c$ using ($ii$).

\noindent(2) By (1) we have $b \rImp a = 0$. Therefore
\begin{align*}
a + a & = a + (a \rImp b)  \tag*{($i$)} \\
	& = b + (b \rImp a) \tag*{$\Cwc$} \\
	& = b.
\end{align*}

\noindent(3) By ($i$) and (1) we have
$ a = a \rImp b \geq a \rImp c $
and hence $0 \geq a \rImp (a \rImp c)$, which implies (3).

\noindent(4) By \Cwc~$x + (x \rImp y) + z = y + (y \rImp x) + z $, whence (4) follows.

\noindent(5) We have
\begin{align*}
c \rImp (b + x) & \geq c \rImp b \\
& = c \tag*{($ii$)}
\end{align*}
and then using (2) we obtain (5).

\noindent(6) 
%
%
By (5), as $(c \rImp a) + a \geq c \rImp (a + a)$, we have $(c \rImp a) + a \geq c $ and hence (6). 

\noindent(7) Our assumptions are symmetric in $a$ and $c$. Hence, (6)
holds with $a$ and $c$ interchanged, i.e., $a \rImp c \geq c \rImp a$,
which taken with (6) gives (7). 

\noindent(8) We have
\begin{align*}
c + (c \rImp a) + ((a \rImp c) \rImp a)
	& = a + (a \rImp c) + ((a \rImp c) \rImp a) \tag*{$\Cwc$} \\
	& = a + a + (a \rImp (a \rImp c)) \tag*{$\Cwc$} \\
	& = b + (a \rImp (a \rImp c)) \tag*{(2)} \\
	& = b. \tag*{(3)}
\end{align*}

\noindent(9) We have
\begin{align*}
b & = c + (c \rImp a) + ((a \rImp c) \rImp a) \tag*{(8)} \\
	& = c + (a \rImp c) + ((a \rImp c) \rImp a) \tag*{(7)} \\
	& = c + a + (a \rImp (a \rImp c)) \tag*{$\Cwc$} \\
	& = c + a. \tag*{(3)}
\end{align*}
This completes the proof of the lemma.
\Done

\vspace{2mm}

\begin{figure}
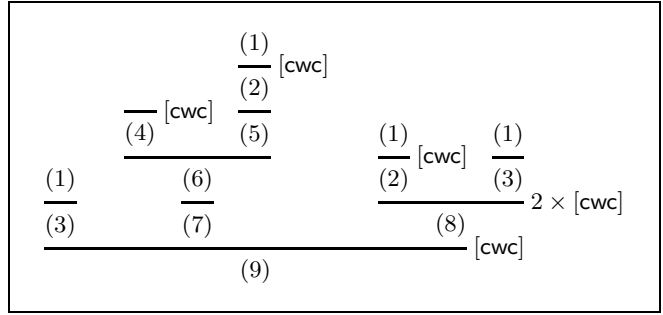

\centering
\begin{tabular}{|ccc|}\hline
\ &&\\
$\quad$ &
\begin{prooftree}
\[
	(1)
	\justifies
	(3)
\]
\quad
\[
	\[
		\[
			\justifies
			(4)
			\using{\Cwc}
		\]
		\[
			\[
				(1)
				\justifies
				(2)
				\using{\Cwc}
			\]
			\justifies
			(5)
		\]
		\justifies
		(6)
	\]
	\justifies
	(7)
\]
\quad
\[
	\[
		(1)
		\justifies
		(2)
		\using{\Cwc}
	\]
	\[
		(1)
		\justifies
		(3)
	\]
	\justifies
	(8)
	\using{2 \times \Cwc}
\]
\justifies
(9)
\using{\Cwc}
\end{prooftree} &
$\quad$
\\\ &&\\\hline
\end{tabular}
\caption{Outline of the Proof of Lemma~\ref{claims}}
\label{fig:outline-proof-tree}
\end{figure}

It is interesting to note the complexity of the proof in terms of uses of $\Cwc$ (used 6 times!) and the important sub-lemma (2) (used twice)
as depicted in the outline proof tree shown in Figure~\ref{fig:outline-proof-tree}.

Finally, from part (9) of Lemma \ref{claims} we have the theorem that
the equation $a \rImp b = a$ uniquely determines $a$ in terms of $b$:

\begin{Theorem} \label{main-thm} In any hoop, if $a \rImp b = a$ and $c \rImp b = c$ then $a = c$.
\end{Theorem}
\Proof Since the assumptions are symmetric in $a$ and $c$ it is enough to show $c \geq a$, from which we can immediately conclude $a \geq c$ and hence $a = c$. By Lemma \ref{claims} (9) we have $c \geq a \rImp b$ and hence $c \geq a$. 
\Done

\vspace{2mm}

We already have the part of Theorem~\ref{thm:ali-sound-complete}  that gives soundness and completeness of $\LLi$ for bounded hoops. Theorem \ref{main-thm} now gives us that the continuous logic axioms $\HL$ and $\HU$ uniquely determine halving given the other axioms of $\LLi$  and that is exactly what we need to complete 
the proof of Theorem~\ref{thm:ali-sound-complete}.

\Section{Subsequent Work}\label{sec:subsequent-work}

The importance of Theorem~\ref{main-thm} is that it provides a powerful method
for proving statements of the form $a = b/2$ in a coop: to prove $a = b/2$, one
proves that $a = a \rImp b$. Very frequently one has to prove statements of the
forms $a \ge b/2$ and $a \le b/2$.  The result on equality suggests that
sufficient conditions for these should be $a \ge a \rImp b$ and $a \le a \rImp
b$ respectively. In logical terms, this means that it is valid to omit either
the first or the last of the antecedents in the inference rule of
Figure~\ref{fig:half-equals-rule}.  Encouraged by our success with
Theorem~\ref{main-thm}, we presented these two problems to Prover9, which, in
just under 4 minutes and just over 20 minutes respectively, found
proofs, that turned out to be even simpler than that of Theorem~\ref{main-thm}.
Once one has these basic tools for reasoning about the halving operator, a
deeper investigation of the algebra of coops becomes possible. One finds for
example, that a coop is simple (in the sense of universal algebra) iff it is
isomorphic to a coop of real numbers under capped addition.
See ~\cite{arthan-oliva12} for more information and for the lovely proofs
found by Prover9 of the rules for $a \ge b/2$ and $a \le b/2$.

Prover9 has also found some other intricate  proofs in
this area. For example, it can prove a lemma on pocrims implying that the axiom
schemata $\CWC + \DN$ is equivalent to $\CSD$ over intuitionistic affine logic
$\ALi$. This implies the aforementioned result that in the $\ALi$-$\LLc$ square
of Figure~\ref{fig:logics}, the north-east logic $\LLc$ is the least extension
of the south-west logic $\ALi$ that contains the other two logics $\ALc$ and
$\LLi$.  Prover9 is able to prove analogous results for each square in
Figure~\ref{fig:logics}. To complement this, Mace4 can also produce the
examples needed to show that the various logics are distinct, with the
exception of the logics in the right-hand column: a non-trivial model of
continuous logic is necessarily infinite and hence not within the scope of
Mace4.

\begin{table}
\begin{center}
\begin{tabular}{|l|l|r|}\hline
TPTP Name & Problem Statement & Seconds\\\hline\hline
LCL888+1.p & Halving is unique: rule for $a = b/2$ & 3.38 \\\hline
LCL889+1.p & Halving is unique: rule for $a \ge b/2$ & 229.13 \\\hline
LCL890+1.p & Halving is unique: rule for $a \le b/2$ ($i$) & 1,216.69 \\\hline
LCL891+1.p & Halving is unique: rule for $a \le b/2$ ($ii$) & 12,724.08 \\\hline
LCL892+1.p & Halving is unique: rule for $a \le b/2$ ($iii$) & 51,876.82 \\\hline
LCL893+1.p & $x/2 = x$ implies $x = 0$ & 0.01 \\\hline
LCL894+1.p & Weak conjunction is l.u.b. in a hoop (Horn) & 1.90  \\\hline
LCL895+1.p & Weak conjunction is l.u.b. in a hoop (Equational) & 14.41 \\\hline
LCL896+1.p & Associativity of weak conjunction implies $\Cwc$ & 5.95 \\\hline
LCL897+1.p & Weak conjunction is associative in a hoop & 0.10 \\\hline
LCL898+1.p & An involutive hoop has $\Csd$ & 66.30 \\\hline
LCL899+1.p & A bounded pocrim with $\Csd$ is involutive & 0.01 \\\hline
LCL900+1.p & A bounded pocrim with $\Csd$ is a hoop & 7.21 \\\hline
LCL901+1.p & An idempotent pocrim with $\Csd$ is boolean & 0.74 \\\hline
LCL902+1.p & A boolean pocrim is involutive & 0.02  \\\hline
LCL903+1.p & A boolean pocrim is idempotent & 1.42 \\\hline
\end{tabular}
\end{center}
\caption{CPU Times for Theorems Contributed to TPTP}
\label{tbl:cpu-times}
\end{table}

A selection of the problems that Prover9 has solved for us will be included
in a forthcoming release of the TPTP Problem Library \cite{Sutcliffe09}.
As can be seen from the CPU times in Table~\ref{tbl:cpu-times},
some of the proof problems are quite challenging. The timings
were taken on an Apple iMac with a 3.06 GHz Intel Core 2 Duo processor
using Prover9's ``auto'' settings. The only tuning we have done is with
the choice of axiomatization. Most of the problems use a straightforward
translation into first-order logic of the various equations and Horn
clauses given above as the axioms for pocrims, hoops etc.
For hoops, a purely equational axiomatization is known
and, in one case (LCL897+1.p), we were unable to obtain a proof using
the Horn axiomatization but obtained a proof very rapidly with the
equational axioms. In other cases (LCL894+1.p, LCL895+1.p), the Horn
axiomatization gives quicker results.

The three axiomatizations we tried for the rule for proving $a \le b/2$
displayed an interesting phenomenon: in the first axiomatization we tried
(LCL890+1.p), we included the annihilator axiom $1 + x = x$, but the
proof, which has 53 steps and was found in about 20 minutes, makes no use of
this. When we tried again without the unnecessary axiom (LCL891+1.p), the
search took an order of magnitude longer and found a proof with 154 steps. When
we put the axiom back in, but this time at the end of the list of axioms
(LCL892+1.p), the search took over 14 hours and gave a proof with 283 steps.
Presumably, in our fortunate first attempt the annihilator axiom had a
beneficial influence on the subsumption process and eliminated a lot of blind
alleys.

When the TPTP formulation of the problems were tried on a selection of
automated theorem provers, only Prover9 was able to find a proof for the first
two problems in less than 300 seconds. Each problem has been proved by at
least one other prover given enough time.  From our perspective as users of this
technology, this is very remarkable: Prover9 delivered a proof of a key lemma
(LCL888+1.p) in just over 3 seconds.  Encouraged by that, we were prepared to
be patient when we tried the two important refinements of that lemma
(LCL889+1.p and LCL890+1.p). These three lemmas have been invaluable in our
subsequent theoretical work on the algebra of coops. We suspect our progress
would have been very different if the first lemma had severely tested our
patience.

\Section{Final Remarks}\label{sec:final-remarks}

We are by no means the first to apply automated theorem proving technology
in the area of {\L}ukasiewicz logics.
In 1990, a conjecture of {\L}ukasiewicz was proposed by Wos
as a challenge problem in automated theorem proving \cite{Wos90}
that was successfully attacked by Anantharaman and Bonacina
\cite{AB90,Bonacina91}.
Others to apply automated theorem proving to {\L}ukasiewicz logics
include Harris and Fitelson \cite{HarrisFitelson01} and Slaney \cite{Slaney02}.
Veroff and Spinks \cite{Veroff-Spinks04} used Otter to find a remarkable direct
algebraic proof of a property of idempotent elements in hoops that had
previously only been proved by indirect model-theoretic methods.



Clearly our application is one to which technology such as Mace4 and Prover9
is well suited.
It is nonetheless a ringing tribute to the late Bill McCune that the
accessibility and ease of use of these tools have enabled two
naive users to get valuable results with very little effort.

\section*{Acknowledgments}
We are grateful to:
the referees for pointers
to the literature and for many other helpful suggestions;
to Roger Bishop Jones for commenting on a draft of the chapter;
to Geoff Sutcliffe for
including our problem set in the TPTP Problem Library and for running the
problems on a selection of provers;
and to Bob Veroff for helping us understand Prover9 performance.
\bibliographystyle{plain}
\bibliography{dhhuh}

\section*{Appendix}

Formal proof of Theorem~\ref{main-thm} as output by Prover9:

\small
\begin{verbatim}
1 x >= y & y >= z -> x >= z # label(non_clause).  [assumption].
2 x >= y & y >= x -> x = y # label(non_clause).  [assumption].
3 x + z >= y <-> z >= x ==> y # label(non_clause).  [assumption].
4 x >= y -> x + z >= y + z # label(non_clause).  [assumption].
5 x >= y -> y ==> z >= x ==> z # label(non_clause).  [assumption].
6 x >= y -> z ==> x >= z ==> y # label(non_clause).  [assumption].
7 y = y ==> x & z = z ==> x -> y = z 
                             # label(non_clause) # label(goal).  [goal].
8 (x + y) + z = x + (y + z).  [assumption].
9 x + y = y + x.  [assumption].
10 x + 0 = x.  [assumption].
11 x >= x.  [assumption].
12 -(x >= y) | -(y >= z) | x >= z.  [clausify(1)].
13 -(x >= y) | -(y >= x) | y = x.  [clausify(2)].
14 -(x + y >= z) | y >= x ==> z.  [clausify(3)].
15 x + y >= z | -(y >= x ==> z).  [clausify(3)].
16 x >= 0.  [assumption].
17 -(x >= y) | x + z >= y + z.  [clausify(4)].
18 -(x >= y) | y ==> z >= x ==> z.  [clausify(5)].
19 -(x >= y) | z ==> x >= z ==> y.  [clausify(6)].
20 x + (x ==> y) = y + (y ==> x).  [assumption].
21 c1 ==> c2 = c1.  [deny(7)].
22 c3 ==> c2 = c3.  [deny(7)].
23 c3 != c1.  [deny(7)].
24 x + (y + z) = y + (x + z).  [para(9(a,1),8(a,1,1)),rewrite([8(2)])].
27 0 + x = x.  [para(10(a,1),9(a,1)),flip(a)].
28 x >= y ==> (y + x).  [hyper(14,a,11,a)].
30 -(x + y >= z) | x >= y ==> z.  [para(9(a,1),14(a,1))].
31 -(x >= y) | 0 >= x ==> y.  [para(10(a,1),14(a,1))].
32 x + (x ==> y) >= y.  [hyper(15,b,11,a)].
33 x >= y ==> 0.  [hyper(14,a,16,a)].
34 x + y >= y.  [hyper(17,a,16,a),rewrite([27(3)])].
35 0 ==> x >= y ==> x.  [hyper(18,a,16,a)].
36 x + ((x ==> y) + z) = y + ((y ==> x) + z).  
                                  [para(20(a,1),8(a,1,1)),rewrite([8(3)])].
41 c3 + x >= c2 | -(x >= c3).  [para(22(a,1),15(b,2))].
43 -(x + (y + z) >= u) | x + z >= y ==> u.  [para(24(a,1),14(a,1))].
46 0 ==> x = x + (x ==> 0).  [para(27(a,1),20(a,1))].
52 x ==> 0 = 0.  [hyper(13,a,16,a,b,33,a),flip(a)].
53 0 ==> x = x.  [back_rewrite(46),rewrite([52(4),10(4)])].
54 x >= y ==> x.  [back_rewrite(35),rewrite([53(2)])].
55 x ==> (y + z) >= x ==> z.  [hyper(19,a,34,a)].
70 x >= y ==> (x + y).  [para(9(a,1),28(a,2,2))].
81 c2 >= c1.  [para(21(a,1),54(a,2))].
82 c2 >= c3.  [para(22(a,1),54(a,2))].
86 x + c2 >= c1.  [hyper(12,a,34,a,b,81,a)].
89 x ==> c2 >= x ==> c3.  [hyper(19,a,82,a)].
127 x >= c2 ==> c1.  [hyper(30,a,86,a)].
171 c2 ==> c1 = 0.  [hyper(13,a,16,a,b,127,a),flip(a)].
180 c1 + c1 = c2.  
         [para(171(a,1),20(a,1,2)),rewrite([9(3),27(3),21(5)]),flip(a)].
205 c1 + (x + c1) = x + c2.  [para(180(a,1),8(a,2,2)),rewrite([9(4)])].
271 x + ((x ==> y) + ((y ==> x) ==> z)) = y + (z + (z ==> (y ==> x))).  
                                        [para(20(a,1),36(a,1,2)),flip(a)].
275 (x ==> y) + z >= x ==> (y + ((y ==> x) + z)).  [para(36(a,1),28(a,2,2))].
418 c1 >= c1 ==> c3.  [para(21(a,1),89(a,1))].
419 0 >= c1 ==> (c1 ==> c3).  [hyper(31,a,418,a)].
609 c3 + (x + (x ==> c3)) >= c2.  [hyper(41,b,32,a)].
895 c3 ==> (x + c2) >= c3.  [para(22(a,1),55(a,2))].
996 c1 ==> (c1 ==> c3) = 0.  [hyper(13,a,16,a,b,419,a),flip(a)].
5220 c3 ==> (c1 + (x + c1)) >= c3.  [para(205(a,2),895(a,1,2))].
10398 c3 + (x ==> c3) >= x ==> c2.  [hyper(43,a,609,a)].
16713 c3 + ((c3 ==> c1) + ((c1 ==> c3) ==> c1)) = c2.  
           [para(996(a,1),271(a,2,2,2)),rewrite([9(15),27(15),180(14)])].
20059 c1 + (c3 ==> c1) >= c3.  [hyper(12,a,275,a,b,5220,a),rewrite([9(5)])].
20066 c3 ==> c1 >= c1 ==> c3.  [hyper(14,a,20059,a)].
20564 c3 + (c1 ==> c3) >= c1.  [para(21(a,1),10398(a,2))].
20570 c1 ==> c3 >= c3 ==> c1.  [hyper(14,a,20564,a)].
20614 c3 ==> c1 = c1 ==> c3.  [hyper(13,a,20066,a,b,20570,a),flip(a)].
20625 c1 + c3 = c2. 
    [back_rewrite(16713),rewrite([20614(4),20(10),996(7),9(4),27(4),9(3)])].
20634 c3 >= c1.  [para(20625(a,1),28(a,2,2)),rewrite([21(4)])].
20637 c1 >= c3.  [para(20625(a,1),70(a,2,2)),rewrite([22(4)])].
20793 -(c1 >= c3).  [ur(13,b,20634,a,c,23,a)].
20794 $F.  [resolve(20793,a,20637,a)].
\end{verbatim}

\end{document}